\documentclass[runningheads]{llncs}
\usepackage{graphicx}
\usepackage{multirow}
\usepackage{tikz}
\usepackage{bbding}
\usepackage{comment}
\usepackage{bbm}
\usepackage{amsmath,amssymb} 
\usepackage{color}
\usepackage[accsupp]{axessibility}  
\usepackage{enumitem}
\usepackage{url}

\begin{document}
\pagestyle{headings}
\mainmatter
\def\ECCVSubNumber{379}  

\title{SdAE: Self-distillated Masked Autoencoder} 

\titlerunning{SdAE: Self-distillated Masked Autoencoder}
%
\author{Yabo Chen$^{1\dagger}$, Yuchen Liu$^{2\dagger}$, Dongsheng Jiang$^{3\dagger}$, Xiaopeng Zhang$^{3}$
, Wenrui Dai$^{1}$, Hongkai Xiong$^{2}$, and Qi Tian$^3$\thanks{Correspondence to Qi Tian. $^{\dagger}$Equal contribution.}}
\authorrunning{Y. Chen et al.}
%
\institute{$^1$Department of Computer Science and Engineering, Shanghai Jiao Tong University\\
$^2$Department of Electronic Engineering, Shanghai Jiao Tong University\\
{\tt\small\{chenyabo, liuyuchen6666, daiwenrui, xionghongkai\}@sjtu.edu.cn}\\
$^3$Huawei Cloud EI\\{\tt\small{dongsheng\_jiang@outlook.com}}, {\tt\small{zxphistory@gmail.com}}, {\tt\small{tian.qi1@huawei.com}} 
}
\maketitle
\begin{abstract}
With the development of generative-based self-supervised learning (SSL) approaches like BeiT and MAE, how to learn good representations by masking random patches of the input image and reconstructing the missing information has grown in concern. 
However, BeiT and PeCo need a “pre-pretraining” stage to produce discrete codebooks for masked patches representing. 
MAE does not require a pre-training codebook process, but setting pixels as reconstruction targets may introduce an optimization gap between pre-training and downstream tasks that good reconstruction quality may not always lead to the high descriptive capability for the model.
Considering the above issues, in this paper, we propose a simple Self-distillated masked AutoEncoder network, namely SdAE.
SdAE consists of a student branch using an encoder-decoder structure to reconstruct the missing information, and a teacher branch producing latent representation of masked tokens.
We also analyze how to build good views for the teacher branch to produce latent representation from the perspective of information bottleneck. After that, we propose a multi-fold masking strategy to provide multiple masked views with balanced information for boosting the performance, which can also reduce the computational complexity.
Our approach generalizes well: with only 300 epochs pre-training, a vanilla ViT-Base model achieves an 84.1\% fine-tuning accuracy on ImageNet-1k classification, 48.6 mIOU on ADE20K segmentation, and 48.9 mAP on COCO detection, which surpasses other methods by a considerable margin. Code is available at \url{https://github.com/AbrahamYabo/SdAE}.
\keywords{Self-supervised Learning, Masked Image Modeling, Vision Transformer}
\end{abstract}

\section{Introduction}
The masked language modeling task (MLM)~\cite{bert2018} has shown great success in self-supervised learning (SSL) for natural language processing.
In computer vision, contrastive learning/instance discrimination~\cite{TingChen2020ASF,JeanBastienGrill2020BYOL,KaimingHe2020MoCo,AaronvandenOord2018RepresentationLW} is a promising direction recently, which regards each instance in the training dataset as a single category. Based on instance discrimination~\cite{MathildeCaron2020UnsupervisedLO,TingChen2020BigSM,XinleiChen2020ImprovedBW}, some methods show the effectiveness in many computer vision tasks.
With the development of vision transformer~\cite{AlexeyDosovitskiy2021ViT}, inspired by natural language processing (NLP), the generative based self-supervised learning (SSL) methods~\cite{baevski2022data2vec,HangboBao2021BEiT,XiaokangChen2022CAE,XiaoyiDong2021PeCo,he2021masked} using masked image modeling (MIM) task have grown in concern. MIM first randomly masks some proportion of image patches, and then recovers the masked patches based on the corrupted image.

\begin{figure}[!t]
\centering
\includegraphics[width=11.8cm]{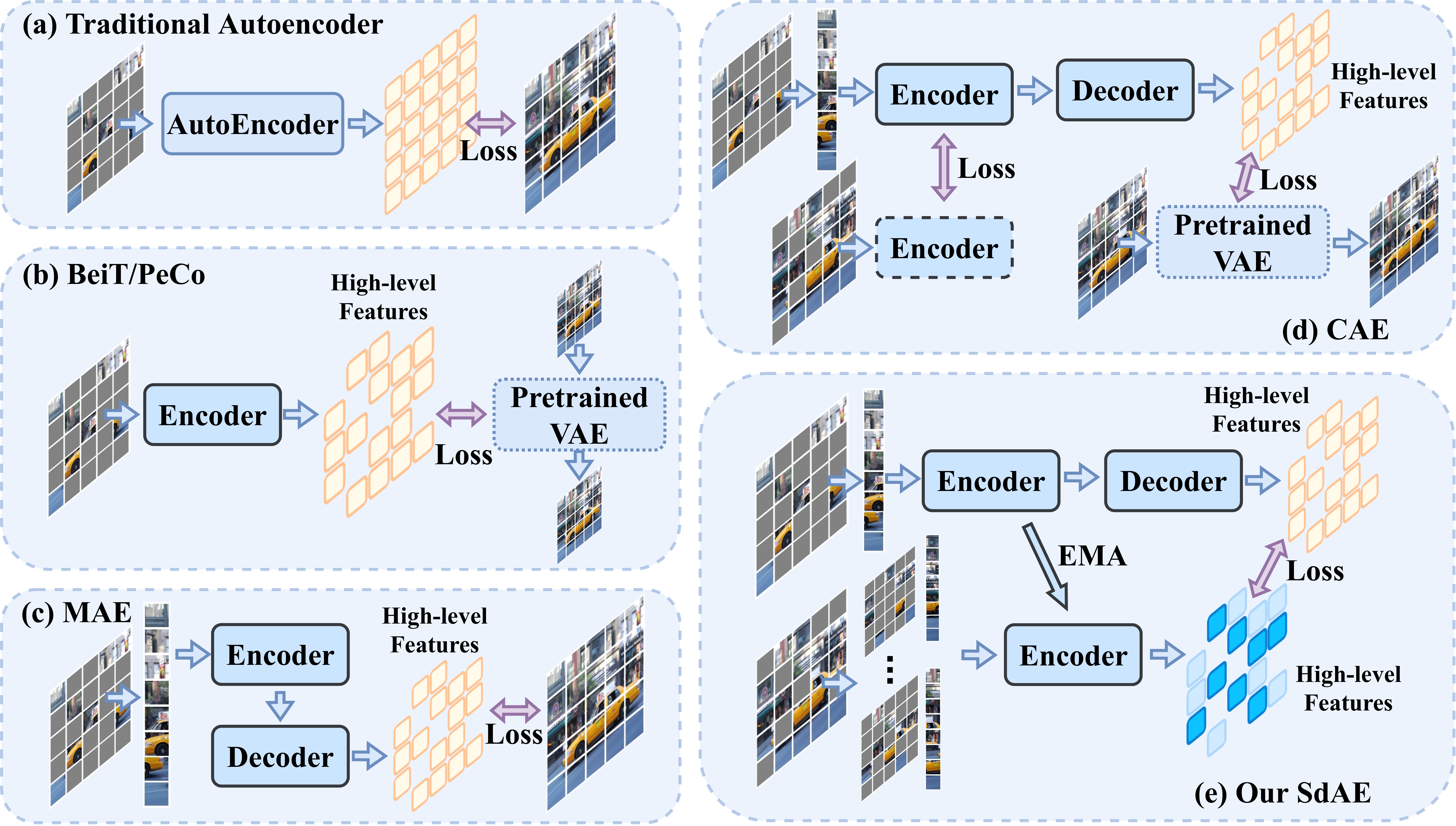}
\caption{
Comparison of typical generative-based self-supervised learning methods.
(a) The inpainting-based methods aim to regress the whole image. (b) BeiT~\cite{HangboBao2021BEiT} and PeCo~\cite{XiaoyiDong2021PeCo} model the discrete visual correlation using a pre-trained VAE. They need a pre-trained stage to get the discrete codebooks. 
(c) MAE~\cite{he2021masked} mainly models the masked image modeling task as a simply pixel restoration task.
(d) Although CAE~\cite{XiaokangChen2022CAE} builds a two-branch structure, it still needs an offline tokenizer. CAE only designs the two-branch to build alignment between masked and unmasked features.
(e) Without the need of a pre-trained codebook, our SdAE can build discrete correlations by self-distillating and constructing the reconstruction targets with high-level features, which can eliminate the representation gaps between
the actual reconstruction targets. 
SdAE also proposes a masking strategy to organize the mask tokens for the teacher branch.
}
\label{fig:compare}
\end{figure}

BeiT~\cite{HangboBao2021BEiT} transfers the MIM task into discrete token classification using a pre-trained discrete autoencoder dVAE (DALL-E~\cite{AdityaRamesh2021dalle}).
PeCO~\cite{XiaoyiDong2021PeCo} modifies the generating procedure of codebook by enforcing the perceptual similarity during the VAE training. 
Similarly, these methods rely on a pre-trained feature descriptor to obtain the latent representation of masked tokens. These designs requiring an additional codebook are a kind of $``$pre-pretraining$"$.

MAE~\cite{he2021masked} proposes an asymmetric encoder-decoder architecture that can reconstruct the raw pixels from the latent representation. MaskFeat~\cite{wei2021masked} uses the hand-crafted feature descriptor Histograms of Oriented Gradients (HOG) to tokenize the image features.
Although these methods do not need additional codebooks, they employ restoring low-level representations such as pixels for masked image modeling tasks. Nevertheless, restoring low-level representations such as pixels is redundant for high semantic level tasks. Moreover, directly reconstructing pixels may lead to an optimization gap between pre-training and downstream tasks, \textit{i.e.}, good reconstruction quality may not always lead to the high descriptive capability of the model. 

Considering the above issue, we propose a simple yet effective self-distillated masked autoencoder structure called SdAE. In SdAE, we claim that MAE itself can produce good representations in an effective and efficient way, and can eliminate the representation gap when used as codebook appropriately.
Without needing a codebook in advance nor modeling a low-level representation, SdAE uses a self-distillated teacher-student network to produce the latent representation as reconstruction targets. 
The student branch consists of the asymmetric encoder-decoder architecture that feeds unmasked images, and the teacher branch contains an encoder to produce latent representation and updates weights from the student using Exponential Moving Average (EMA).

When introducing the teacher branch into the masked autoencoder structure, the easiest way is to feed the full image into the teacher network directly.
However, there is no computational loss on the unmasked tokens. Obviously, due to the spatial redundancy that exists in the image, it is not optimal to put the whole image into the teacher branch.
In addition, simply putting all masked tokens is still resource-consuming and faced with performance degradation compared with using the whole image as input.
So there grows another concern about how to produce better latent representation using the raw images.

We further discuss this problem from the perspective of information bottleneck and propose a multi-fold masking strategy to produce good views for the self-distillated masked autoencoders as well as reduce the computational complexity.
The contributions of this paper are summarized as below: 
\begin{itemize}[itemsep=1pt,topsep=1pt]
\item We propose a novel self-distillated masked autoencoder structure that can construct a learnable high-level reconstruction target rather than extra pre-trained codebooks or low-level pixels, and find that MAE itself can produce a better codebook.
\item We discuss how to produce good views for the teacher branch and propose a multi-fold masking strategy to keep mutual information from the teacher branch relevant to the student one. This strategy can also save computation resources.
\end{itemize}
\section{Methods}
\begin{figure}[!t]
\centering
\includegraphics[width=11.8cm]{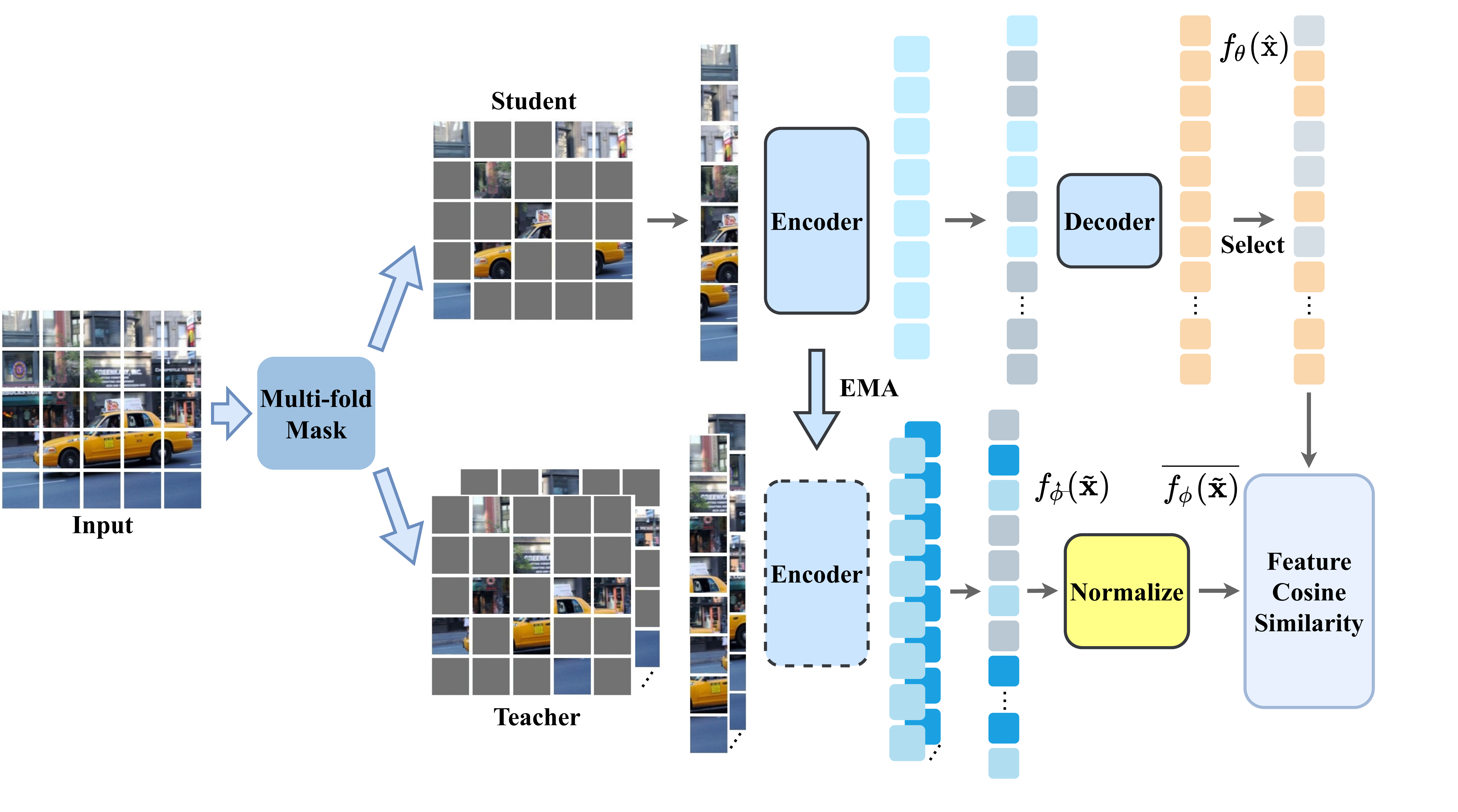}
\caption{An overall framework of SdAE. SdAE adopts ViT~\cite{AlexeyDosovitskiy2021ViT} as the backbone and consists of two branches. The student branch with an asymmetric encoder-decoder reconstructs the missing patches, referring to the teacher branch's latent representation. The teacher branch is simply an exponential moving average (EMA) of the student one. 
Specifically, all masked tokens are divided into several $``$small parts$"$ to fit the information bottleneck between two branches for the multi-fold masking strategy.
}
\label{fig:architecture}
\end{figure}

Firstly, we elaborate the basic framework of masked image modeling and derive the objective function of our masked feature reconstruction methods in Section~\ref{ssec:framework}.  \figurename~\ref{fig:architecture} depicts our proposed self-distillated masked autoencoder that consists of a two-branch network, \textit{i.e.}, teacher and student branches.
After that, we present theoretical discussions on how to produce good views for the teacher branch to build latent representation. In addition, we propose a multi-fold masking strategy, which is tailored for balancing the information between the teacher and student branches as in Section~\ref{ssec:multifold}. 
Finally, in Section~\ref{ssec:distill}, the distillation strategy for the teacher-student framework is demonstrated.

\subsection{Framework}\label{ssec:framework}
Inspired by Masked Language Modeling (MLM) as the pre-training tasks in BERT~\cite{bert2018}, Masked Image Modeling (MIM) has been proposed in several recent works~\cite{HangboBao2021BEiT,he2021masked,xie2021simmim}. Specifically, an image $\mathbf{x} \in \mathbb{R}^{H \times W \times C}$ is reshaped and tokenized into a sequence of flattened 2D patches as $\mathbf{x}=\left\{\mathbf{x}_{i}\right\}_{i=1}^{N}$, where $\mathbf{x}_{i} \in \mathbb{R}^{N \times\left(P^{2} \cdot C\right)}$ and $(P,P)$ is the resolution of each image patch. 
MIM firstly samples a random binary mask $\mathbf{M} \in \{0, 1\}^{N}$ to be $\left\{m^{1}, m^{2}, \cdots, m^{N}\right\}$, where $m^{i}\in\{0,1\}$ and $N$ is the number of tokens. The patch token $\mathbf{x}_i$ with $m^i=1$ is masked and we denote the mask ratio as $\mathrm{r}=\frac{\sum_{i=1}^{N}m^i}{N} \in (0,1) $. All the masked tokens $\tilde{\mathbf{x}} \triangleq\left\{\mathbf{x}_{i} \mid m^{i}=1\right\}_{i=1}^{N}$ are discarded for efficiency, resulting in a corrupted image $\hat{\mathbf{x}} \triangleq\left\{\hat{\mathbf{x}}_{i} \mid\left(1-m^{i}\right) \mathbf{x}_{i}\right\}_{i=1}^{N}$. In summarize, we have $\tilde{\mathbf{x}} \in \mathbb{R}^{ N \cdot \mathrm{r} \times\left( P^{2} \cdot C\right)}$ and $\hat{\mathbf{x}} \in \mathbb{R}^{ N \cdot (1-\mathrm{r}) \times\left( P^{2} \cdot C\right)}$. The objective of MIM is to recover the masked tokens $\tilde{\mathbf{x}}$ by using the corrupted image $\hat{\mathbf{x}}$ as $\log q_{\mathbf{\psi}}(\tilde{\mathbf{x}}|\hat{\mathbf{x}})$. Assuming the reconstruction of each masked token is independent to each other~\cite{HangboBao2021BEiT,JinghaoZhou2021iBOT}, the objective can be reformulated as:
\begin{equation}
    \log q_{\mathbf{\psi}}(\tilde{\mathbf{x}}| \hat{\mathbf{x}}) \approx \sum_{i=1}^{N} m^{i} \cdot \log q_{\mathbf{\psi}}\left(\mathbf{x}_{i}| \hat{\mathbf{x}}\right)
    \label{eq:mim}
\end{equation}

Considering MIM is essentially a reconstruction task. It is related more to regression tasks than to classification tasks. 
Thus, we assume the noise in the deviation of reconstructed and ground truth values follow the standard Gaussian distribution as $\eta \sim N(0,1)$, 
and the objective is to minimize
\begin{equation}
    \sum_{i=1}^{N} m^{i} \cdot \log q_{\mathbf{\psi}}\left(\mathbf{x}_{i}| \hat{\mathbf{x}}\right)=\sum_{i=1}^{N} m^{i} \cdot \log e^{\frac{-(f_{\phi}(\mathbf{x}_i)-f_{\theta}(\hat{\mathbf{x}}))^2}{2}}=\sum_{i=1}^{N}-\frac{1}{2}m^{i}(f_{\phi}(\mathbf{x}_i)-f_{\theta}(\hat{\mathbf{x}}))^2.
    \label{eq:mse}
\end{equation}
Specifically, for MAE~\cite{he2021masked}, $f_{\phi}$ is the identity function, and $f_{\theta}$ is the masked autoencoder that reconstructs masked patches in the pixel spatial space. However, we argue that 
(1) The $f_{\phi}$ is a fixed identity function without adaptation during pre-training that undermines the effectiveness of the self-supervised training;
(2) Memorying each pixel of the images by reconstruction in the low semantic level space is sub-optimal and less efficient for capturing representations for high-level tasks. There exists an optimization direction gap in that the quality of the reconstruction may not always increase the descriptive capability of the model.

Correspondingly, we introduce a self-distillation architecture, \textit{i.e.}, $f_{\theta}$ is the student network trained by gradient descent using $\hat{\mathbf{x}}$ as inputs and $f_{\phi}$ is the teacher network updated by exponential moving average (EMA) from the parameters of the student network. The teacher network use $\tilde{\mathbf{x}}$ as inputs and conduct the missing information reconstruction in the high-level latent representation space rather than in the pixel space.  

Inspired by MAE~\cite{he2021masked}, we propose a value normalization function upon teacher outputs. Specifically, we compute the mean and standard deviation of feature values within a patch and use them to normalize the teacher outputs as
\begin{equation}
\overline{f_{\phi}(\mathbf{x}_i)} = \frac{f_{\phi}(\mathbf{x}_i) - mean(f_{\phi}(\mathbf{x}_i))}{\sqrt{var(f_{\phi}(\mathbf{x}_i)) + \epsilon}}
\end{equation}
where $\epsilon$ is a small value to prevent the denominator from being 0. We find that using normalized features as the reconstruction target improves the representation quality.

Then we minimize the normalized teacher features with the output features of the student decoder based on the feature cosine similarity, and Eq.~\eqref{eq:mse} is reformulated as:
\begin{equation}
\log q_{\mathbf{\psi}}(\tilde{\mathbf{x}}| \hat{\mathbf{x}}) \approx \frac{\sum_{i=1}^{n} m^{i} \overline{f_{\phi}(\mathbf{x}_{i})} f_{\theta}(\hat{\mathbf{x}})}{\sqrt{\sum_{i=1}^{n} m^{i}\left(\overline{f_{\phi}(\mathbf{x}_{i})}\right)^{2}} \sqrt{\sum_{i=1}^{n} m^{i}\left(f_{\theta}(\hat{\mathbf{x}})\right)^{2}}}
\label{eq:sae_mae}
\end{equation}

\subsection{Discussions on The Teacher Branch Feeding}\label{ssec:multifold}
As mentioned in MAE~\cite{he2021masked}, languages are human-generated signals that are highly semantic and information-dense.
Predicting a few missing words per sentence can induce sophisticated language understanding, but images are natural signals with heavy spatial redundancy. 
Directly feeding the whole image into the teacher network to produce features may be sub-optimal.

\begin{figure}[!t]
\centering
\includegraphics[width=8.8cm]{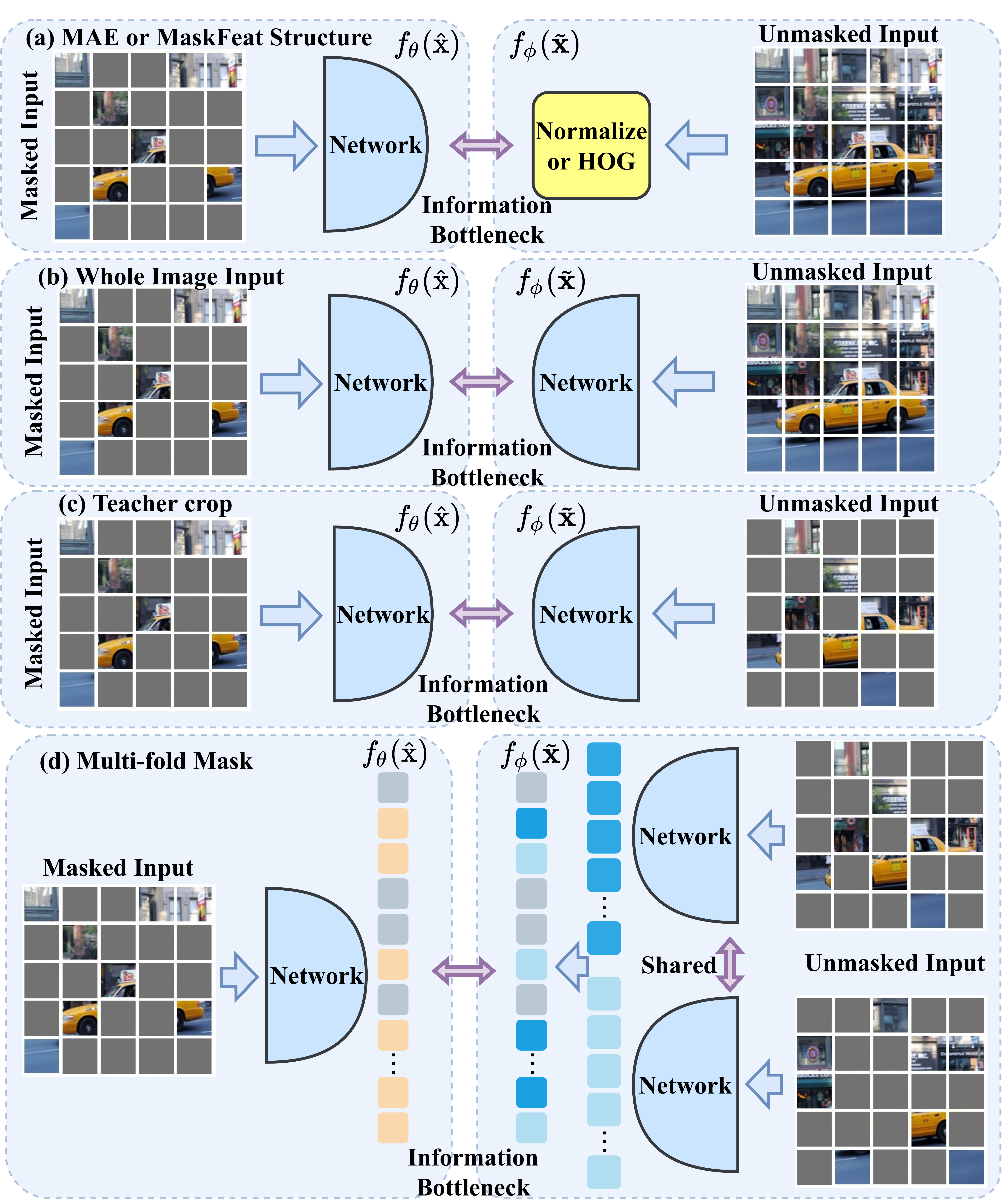}
\caption{(a) MAE~\cite{he2021masked} and MaskFeat~\cite{wei2021masked} reconstruct the low-level representations. (b) Feeding the whole image into the teacher network is the simplest way. (c) We experiment with teacher-crop reconstructing only parts of tokens. (d) Multi-fold masking divides unmasked tokens into independent groups and feeds them into shared networks.}
\label{fig:multifold}
\end{figure}

Before discussing the better feeding of the teacher branch, we provide a brief introduction here to show alternatives to the teacher branch feeding. As shown in \figurename~\ref{fig:multifold} (a), MAE~\cite{he2021masked} and MaskFeat~\cite{wei2021masked} take the whole image as input to produce normalized tokens (normalized raw image pixels) or HOG from the whole images as the reconstruction target. 
Correspondingly, another simply modeling way is to feed the whole image into the teacher network directly as \figurename~\ref{fig:multifold} (b).
However, since we have mentioned that the reconstruction loss only computes for masked tokens, there is no need to feed all the patches, and visible patches may be spatial redundant. Only feeding the masked patches $\hat{\mathbf{x}}$ into the network can relieve the spatial redundancy to some extent. 
Furthermore, to reduce more spatial redundancy and save the computational resource, we can mask some of the target tokens and feed the remaining tokens into the teacher. We called this operation as teacher crop. In the experiment, we can achieve comparable performance between teacher crop and reconstruct all masked patches. As shown in \figurename~\ref{fig:multifold} (c). 
We formulate this teacher crop masking as another random binary mask $\mathbf{M^t} \in \{0, 1\}^{  N \cdot \mathrm{r} }$ to be $\left\{m_t^{1}, m_t^{2}, \cdots, m_t^{  N \cdot \mathrm{r} }\right\}$, where $m_t^{i}\in\{0,1\}$ and $ N \cdot \mathrm{r} $ is the number of total target tokens. Similarly, we formulate the teacher crop mask ratio as $\mathrm{r}_{c} \in (0,1) $. Based on this binary mask, we can randomly select a group of target tokens as: $\tilde{\mathbf{x}}^{\ast} \triangleq\left\{\tilde{\mathbf{x}}_{i} \mid m_t^{i}=1\right\}_{i=1}^{ N \cdot \mathrm{r}}$, and the latent representation is denoted as $f_{\theta}(\tilde{\mathbf{x}}^{\ast})$. 

\textbf{Multi-fold Masking Strategy.} 
Furthermore, we propose a multi-fold masking strategy that groups several teacher crops to sufficiently use masked tokens. As shown in \figurename~\ref{fig:multifold} (d), given a number of fold $t$, 
all the masked tokens are divided into $t$ groups of teacher crops without overlap as $\tilde{\mathbf{x}} = \{{\tilde{\mathbf{x}}^{\ast1}}, {\tilde{\mathbf{x}}^{\ast2}}, ..., {\tilde{\mathbf{x}}^{\ast t}}\} $. Each fold containing masked tokens will be fed into a shared teacher network independently. The outputs of the teacher network will be rearranged together to calculate the cosine similarity loss in a parallel way. Specifically, the output of the teacher branch can be reformulated as:
$f_{\theta}(\tilde{\mathbf{x}}) =\{f_{\theta}({\tilde{\mathbf{x}}^{\ast 1}}),f_{\theta}({\tilde{\mathbf{x}}^{\ast 2}}),...,f_{\theta}({\tilde{\mathbf{x}}^{\ast t}})\}$.
As shown in Table~\ref{tab:complexity}, from the view of computational cost, the main difference between multi-fold masking strategy and all masked patches feeding is that in the teacher branch, the self-attention is only calculated in each teacher crop of tokens instead of all masked tokens. 
\begin{table}[!t]
\renewcommand{\baselinestretch}{1.0}
\renewcommand{\arraystretch}{1.0}   
\begin{center}
\caption{Alternatives for the input of the teacher branch network. The multi-mask strategy captures content only within each fold at a reduced memory cost compared with all masked patches fed. 
Comparatively, using a multi-mask strategy reduces complexities by a factor of $t$. $n$: input length, $d$: dimension of channels, $\mathrm{r}$: mask ratio, $\mathrm{r}_{c}$ the teacher crop mask ratio, $t$: number of fold. The training time is the computational costs for per epoch with NVIDIA V100 GPUs and the memory cost is the maximum usage on per GPU.}\label{tab:complexity} 
\begin{tabular}{c|c|c|c}
\hline 
Operation & Theoretical complexity & Training Time &GPU Memory\\
\hline 
MAE(baseline)&  - & 551s & 14.93G\\
Full image input&  $\Theta(n^2 d)$ & 826s(\textcolor{blue}{+49.9\%}) & 16.24G\\
Only masked patches input&  $\Theta(n^2 r^2 d)$  & 761s(\textcolor{blue}{+38.1\%}) & 16.01G\\
Teacher crop input& $\Theta( n^2 {\mathrm{r}_{c}}^2 \mathrm{r}^2 d)$ & 616s(\textcolor{blue}{+11.8\%}) & 15.54G\\
Multi-fold masking strategy& $\Theta( (n / t)^2 t r^2 d )$  & 706s(\textcolor{blue}{+28.1\%}) & 16.02G\\
\hline
\end{tabular}
\end{center}
\end{table}
So the multi-mask strategy can save complexities by a factor of $t$.  Detailed theoretical complexity can be seen in Table~\ref{tab:complexity}.

In practice, directly feeding the whole image into the teacher network will increase almost half of the computation costs compared with MAE, as shown in Table~\ref{tab:complexity}. Using all masked patches as input like CAE~\cite{XiaokangChen2022CAE} will also increase $38.1\%$ pre-training time. Using multi-fold masking strategy, we can save the extra pre-training time costs by only $28.1\%$. 

The main difference between our multi-fold masking and previous work multi-crop~\cite{MathildeCaron2020UnsupervisedLO} is discussed as follows. 
The multi-crop strategy needs to train multiple random views with different sizes without concern of complexity. 
By contrast, multi-fold masking just rearranges the mask tokens and does not create new views. Multi-fold masking can also save training complexity by calculating self-attention on only a small group of tokens.

\textbf{Information Bottleneck and Multi-fold masking}\label{ssec:bottleneck} 
In this section, we will follow the assumption from the views of information bottleneck theory in \cite{AlexanderAAlemi2016DeepVI}\cite{tian2020makes} to discuss the benefits of multi-fold masking except for saving costs.

Denote $\tilde{\mathbf{x}}$ and $\hat{\mathbf{x}}$ as the input of the teacher and student network, respectively.
From the perspective of information theory, maximizing Eq.~\eqref{eq:mse} is equivalent to maximizing the mutual information $I(\tilde{\mathbf{x}},\hat{\mathbf{x}})$ as
\begin{equation}
    I(\tilde{\mathbf{x}},\hat{\mathbf{x}})=H(f_{\phi}(\tilde{\mathbf{x}}))-H(f_{\phi}({\tilde{\mathbf{x}}})|f_{\theta}({\hat{\mathbf{x}}}))=H(f_{\phi}(\tilde{\mathbf{x}}))+\mathbb{E}_{\tilde{\mathbf{x}},\hat{\mathbf{x}}}[\log p(f_{\phi}(\tilde{\mathbf{x}})|f_{\theta}(\hat{\mathbf{x}}))]
    \label{eq:mutual}
\end{equation}
Since the true distribution of $p(\tilde{\mathbf{x}}|\hat{\mathbf{x}})$ is unknown, a variational distribution $q(\tilde{\mathbf{x}}|\hat{\mathbf{x}})$ is employed for approximation, and Eq.~\eqref{eq:mutual} can be written as:
\begin{equation}
    I(\tilde{\mathbf{x}},\hat{\mathbf{x}})=H(f_{\phi}(\tilde{\mathbf{x}}))+\mathbb{E}_{\tilde{\mathbf{x}},\hat{\mathbf{x}}}[\log q(f_{\phi}(\tilde{\mathbf{x}})|f_{\theta}(\hat{\mathbf{x}}))]+\mathbb{E}_{\hat{\mathbf{x}}}[KL(p(f_{\phi}(\tilde{\mathbf{x}})|f_{\theta}(\hat{\mathbf{x}})||q(f_{\phi}(\tilde{\mathbf{x}})|f_{\theta}(\hat{\mathbf{x}}))]
    \label{eq:variational}
\end{equation}
where $f_{\phi}(\tilde{\mathbf{x}})$ and $f_{\theta}(\hat{\mathbf{x}})$ are the representations from the teacher and student networks, and $H(\cdot)$ is the entropy function. Since $KL(\cdot)$ function is non-negative and $H(\tilde{\mathbf{x}})$ is constant w.r.t the parameters to be optimized, maximizing $I(\tilde{\mathbf{x}},\hat{\mathbf{x}})$ is equivalent to maximizing $\mathbb{E}_{\tilde{\mathbf{x}},\hat{\mathbf{x}}}[\log q(f_{\phi}(\tilde{\mathbf{x}})|f_{\theta}(\hat{\mathbf{x}}))]$ with $q$ modeled by a Gaussian distribution, which is in consistent with Eq.~\eqref{eq:mse}. Thus, the objective of MIM can be reinterpreted as $I(\tilde{\mathbf{x}},\hat{\mathbf{x}})$ in the information theory view.
\\
\textbf{Definition 1.} \emph{(Sufficient Encoder) The encoder $f_{\theta}$ of $\hat{\mathbf{x}}$ is sufficient in the self-supervised learning framework if and only if $\left| I(\tilde{\mathbf{x}},\hat{\mathbf{x}}) - I(\tilde{\mathbf{x}},f_{\theta}(\hat{\mathbf{x}})) \right| \textless \epsilon$.}

Ideally, the encoder $f_{\theta}$ is sufficient if the amount of information in $\hat{\mathbf{x}}$ about $\tilde{\mathbf{x}}$ is lossless during
the encoding procedure. In other words, $f_{\theta}(\hat{\mathbf{x}})$ has kept all the information that the reconstruction objective requires. Symmetrically, $f_{\phi}$ is sufficient if $\left| I(\tilde{\mathbf{x}},\hat{\mathbf{x}}) - I(\tilde{\mathbf{x}},f_{\theta}(\hat{\mathbf{x}})) \right| \textless \epsilon$.\\
\textbf{Definition 2.} \emph{(Minimal Sufficient Encoder) A sufficient encoder $f_{\theta}$ of $\hat{\mathbf{x}}$ is minimal if and only if $I(f_{\theta}(\hat{\mathbf{x}}),\hat{\mathbf{x}}) \leq I(f(\hat{\mathbf{x}}),\hat{\mathbf{x}})$, $\forall f$ that is sufficient.}

Among these sufficient encoders, the minimal ones only keep relevant information of the reconstruction task and throw away other irrelevant information.\\
\textbf{Proposition 1.} \emph{Suppose $f_{\theta}$ and $f_{\phi}$ are minimal sufficient encoders. Given a downstream task $\mathcal{T}$ with label $\mathbf{y}$, the optimal views created from the data $\mathbf{x}$ are $(\tilde{\mathbf{x}}^{*},{\hat{\mathbf{x}}}^{*})=\arg \min _{\tilde{\mathbf{x}}, \hat{\mathbf{x}}} I\left(\tilde{\mathbf{x}}, \hat{\mathbf{x}}\right)$, subject to $\left| I(\tilde{\mathbf{x}},\mathbf{y})-I(\hat{\mathbf{x}},\mathbf{y}) \right| \textless \epsilon $}. 

Specifically, based on Proposition 1, we can model the multi-fold masking tokens as multi-views from reconstructed images, which can be formulated as $\{{\tilde{\mathbf{x}}^{\ast 1}}, {\tilde{\mathbf{x}}^{\ast 2}}, ..., {\tilde{\mathbf{x}}^{\ast t}}\} $ and the unmasked tokens is another view of input images $\hat{\mathbf{x}}$. Then, we draw insights for views that feed into the teacher and student networks: 
(1) Ours method satisfies with $\arg \min _{\tilde{\mathbf{x}}, \hat{\mathbf{x}}} I\left(\tilde{\mathbf{x}}, \hat{\mathbf{x}}\right)$, since the feeding of the teacher and student networks do not contains overlap tokens, reducing the shared mutual information.
(2) Due to the need to satisfy $\left| I(\tilde{\mathbf{x}},\mathbf{y})-I(\hat{\mathbf{x}},\mathbf{y}) \right| \textless \epsilon $, the amount of mutual information between $\tilde{\mathbf{x}}$ and $\hat{\mathbf{x}}$ should be comparable. We propose a multi-fold masking strategy to divide the masked patches into groups to make the number of masked patches feeding into the teacher network equal with that feeding into the student. Thus, their shared mutual information with the downstream task is comparable.
As shown in \figurename~\ref{fig:ema} (b), we experimentally verify this proposition that when the number of masked patches from each fold is equal with the unmasked patches (25\% of total image patches in practice), our SdAE achieves the best performance.
(3) Considering the objective function of MIM tasks is $\max I(\tilde{\mathbf{x}},\mathbf{y})$, we also need to $\max I(\hat{\mathbf{x}},\mathbf{y})$, so we try to utilize almost all masked tokens to keep as more task related information as possible.

\subsection{Distillation Strategy}\label{ssec:distill}
Considering a teacher network, we hypothesize that it is desirable to build the reconstruction target representations in the (i) online and (ii) consistent way. Now that the student network $f_{\theta}$ is trained by back-propagation to minimize the feature reconstruction loss. The teacher network is updated in a momentum update way using exponential moving average (EMA).
Specifically, denoting the parameters of $f_{\phi}$ as $\phi$ and those of $f_{\theta}$ as $\theta$, we update $\phi$ by:
\begin{equation}
    \phi \leftarrow \eta\cdot\phi+(1-\eta)\cdot\theta
\label{eq:ema}
\end{equation}
Here $\eta \in[0,1)$ is a momentum coefficient to control the frequency of updates from the student model. The codebook should not update too often. Otherwise, the model may fail to converge.

\section{Experiments}
This section evaluates our pre-trained feature representation on several unsupervised benchmarks. We first evaluate the classification performance on ImageNet-1k under fine-tuning and linear probing. Then we transfer the pre-trained features to several downstream tasks, \textit{i.e.}, semantic segmentation and object detection. Finally, we conducted an ablation study on the key components of SdAE.
\subsection{Fine-tuning on ImageNet-1k}
We study the fine-tuning on the ILSVRC-2012 ImageNet dataset~\cite{russakovsky2015imagenet} with 1k classes and 1.3M images. For a fair comparison, we directly follow most of the hyperparameters of MAE~\cite{he2021masked} in our fine-tuning experiments. All experiments reported are
only fine-tuning for 100 epochs (vs. 300 training from scratch).
We compare our SdAE with Vision Transformers trained by random initialization and previous self-supervised learning methods.
As shown in Table~\ref{tab:classify}, compared with the models trained by random initialization which only achieves 81.8\% top-1 accuracy with ViT-B, our SdAE achieves 84.1\%, demonstrating the effectiveness of pre-training with unlabeled data.

Compared with previous self-supervised methods, our proposed SdAE surpasses them on ImageNet fine-tuning by a large margin. For ViT-B, our SdAE outperforms MAE by 1.2\% top-1 accuracy with the same number of training epochs, demonstrating that MIM on high-level latent feature space is more effective than low-level pixel space. 
Besides, our SdAE outperforms BEiT by 1.1\% top-1 accuracy. Moreover, compared to the recently proposed CAE, our SdAE achieves 0.8\% top-1 accuracy gain, demonstrating the effectiveness of our self-distillated design and multi-fold masking strategy. In addition, with only 100 epochs pre-training, SdAE can achieve comparable performance with MAE using 1600 epochs pre-training and surpass 300 epochs pre-trained CAE. Our proposed SdAE also surpasses above methods on ImageNet linear probing with the same training epochs. As a MIM based method, SdAE can also surpasses MIM based methods with the same pre-training epochs. The phenomenon that contrastive based methods surpass the MIM based ones on linear probing is also discussed in MAE~\cite{he2021masked} and CAE~\cite{XiaokangChen2022CAE}. In terms of linear probing, contrastive learning mainly cares about the 1000 classes and MIM methods may care about the classes beyond the 1000 classes. So fine-tuning measurement may better validate the effectiveness of MIM based methods.
\begin{table}[!t]
\renewcommand{\baselinestretch}{1.0}
\renewcommand{\arraystretch}{1.0}
\setlength{\tabcolsep}{6pt}
\begin{center}
\caption{Image classification results on the ILSVRC-2012 ImageNet dataset with top 1 accuracy. $``$Epochs$"$ refers to the number of pre-training epochs. MoCo v3 and DINO
adopt multi-crop augmentation for pre-training. MoCo v3: 2 global crops of 224 × 224. DINO: 2 global crops of
224 × 224 and 10 local crops of 96 × 96.}\label{tab:classify} 
\begin{tabular}{lccccc}
\hline Method & Epochs & Crops & Finetune & Linear\\
\hline \emph{Methods using ViT-B}: & &   \\
Train from Scratch&  $300$ & $-$ & 81.8 & $-$   \\
MoCo v3 & 300 & 2 & $83.2$ & $76.2$  \\
DINO & 400 & 12 & $83.3$ &$\textbf{77.3}$\\
BEiT & 300 & 1 & $83.0$ & $49.4$ \\
MAE & 100 & 1 & $82.1$ & $54.8$\\
MAE & 300 & 1 & $82.9$ & $61.5$\\
MAE & 1600 & 1 & $83.6$ & $67.8$ \\
CAE & 300 & 1 & $83.3$ & $64.2$ \\
SdAE & 100 & 1 & $83.5$ &  $60.3$ \\
SdAE & 300 & 1 & $\textbf{84.1}$ & $64.9$ \\
\hline
\end{tabular}
\end{center}
\end{table}

\subsection{Semantic Segmentation}
We evaluate the learned representation of our SdAE on the ADE20K benchmark~\cite{zhou2019semantic} with 25K images and 150 semantic categories. The mean Intersection of Union (mIoU) averaged over all semantic categories is reported as the evaluation metric. Table~\ref{tab:seg} shows that our SdAE achieves the state-of-the-art performance with 48.6 mIoU for 300 pre-training epochs. Our SdAE outperforms BEiT, MAE, and CAE by 3.1, 2.8, and 1.9 mIoU, respectively. Besides, our SdAE with 300 pre-training epochs even outperforms MAE with 1600 pre-training epochs.
\begin{table}[!t]
\renewcommand{\baselinestretch}{1.0}
\renewcommand{\arraystretch}{1.0}
\setlength{\tabcolsep}{6pt}
\begin{center}
\caption{Semantic segmentation on ADE20K. All methods use ViT-B backbone based on the same implementation. $``$Epochs$"$ refers to the number of pre-training epochs.}\label{tab:seg} 
\begin{tabular}{lccccc}
\hline Method & Epochs & Crops& Supervised & Self-supervised & mIoU \\
\hline 
DeiT &300& $-$ & $\checkmark$ & $\times$ & $47.0$ \\
MoCo v3 &300 & 2 & $\times$ & $\checkmark$ & $47.2$ \\
DINO & 400 & 12  & $\times$ & $\checkmark$  & $47.2$ \\
BEiT & 300 & 1 & $\times$ & $\checkmark$  & $45.5$ \\
BEiT & 800 & 1 & $\times$ & $\checkmark$ &  $46.5$ \\
MAE & 300 & 1 & $\times$ & $\checkmark$ & $45.8$ \\
MAE & 1600 & 1 & $\times$ & $\checkmark$ & $48.1$ \\
CAE & 300 & 1 & $\times$ & $\checkmark$ & $47.7$ \\
SdAE & 300 & 1 & $\times$ & $\checkmark$ & $\textbf{48.6}$ \\
\hline 
\end{tabular}
\end{center}
\end{table}

\subsection{Object Detection}
Following CAE~\cite{XiaokangChen2022CAE}, we fine-tune Mask R-CNN~\cite{he2017mask} in an end-to-end manner on COCO~\cite{lin2014microsoft}. The ViT backbone is adapted for use with FPN~\cite{lin2017feature}. The box AP for object detection and the mask AP
for instance segmentation is reported in Table~\ref{tab:obj}. Our method
(300 epochs, ViT-B) is consistently superior to all the other models. Our SdAE performs better than the recent published CAE (48.9 \emph{vs.} 48.0 AP$^b$). Besides, it is worth mentioning that our SdAE (300 epochs) even outperforms MAE (1600 epochs) by 0.5 AP$^b$. As an effective framework for self-supervised learning, we achieve better performance with fewer training epochs.

\begin{table}[!tbp]
\renewcommand{\baselinestretch}{1.0}
\renewcommand{\arraystretch}{1.0}
\setlength{\tabcolsep}{3pt}
\begin{center}
\caption{Object detection and instance segmentation on COCO. All the results are based on the same implementations. Mask R-CNN is adopted and trained with the 1× schedule.  $``$Epochs$"$ refers to the number of pre-training epochs on ImageNet-1K. $``$Sup$"$ and $``$Self-sup$"$ refer to the methods that are supervised or self-supervised. 
}\label{tab:obj} 
\begin{tabular}{lcccp{22pt}p{22pt}p{22pt}p{23pt}p{23pt}p{23pt}}
\hline \multirow{2}*{Methods} & \multirow{2}*{Epochs} &  \multirow{2}*{Sup} &  \multirow{2}*{Self-sup}&\multicolumn{3}{c}{Object Detection}&\multicolumn{3}{c}{Instance Segmentation}\\
\cline{5-10}
&&&&AP$^b$&AP$_{50}^b$&AP$_{75}^b$&AP$^m$&AP$_{50}^m$&AP$_{75}^m$\\
\hline
\multicolumn{3}{l}{\emph{Methods using ViT-B:}}\\
DeiT & 300 & $\checkmark$ & $\times$ &46.9&68.9& 51.0& 41.5& 65.5& 44.4\\
MoCo v3 & 300 & $\times$ & $\checkmark$ &45.5 &67.1 &49.4& 40.5& 63.7& 43.4\\
DINO&400& $\times$ & $\checkmark$ &46.8 &68.6& 50.9& 41.5& 65.3& 44.5\\
BEiT&300& $\times$ & $\checkmark$ &39.5 &60.6 &43.0 &35.9& 57.7& 38.5\\
BEiT&800& $\times$ & $\checkmark$ &42.1 &63.3& 46.0& 37.8& 60.1& 40.6\\
MAE & 300 & $\times$ & $\checkmark$ & 45.4 &66.4& 49.6& 40.6& 63.4& 43.7 \\
MAE & 1600 & $\times$ & $\checkmark$ & 48.4 &69.4& 53.1& 42.6& 66.1& 45.9 \\
CAE & 300 & $\times$ & $\checkmark$ & 48.0& 68.7& 52.7& 42.3 &65.6& 45.4 \\
SdAE & 300 & $\times$ & $\checkmark$ & \textbf{48.9} & 69.6 & 53.3 & \textbf{43.0} & 66.2 & 46.2 \\
\hline 
\end{tabular}
\end{center}
\end{table}


\section{Ablation Studies}
 In this section, we present ablation studies to better evaluate the contributions of each component and hyperparameter settings in our proposed SdAE. Unless specified, all results are compared with
models pre-trained for 100 epochs for efficiency, and we report the top-1 accuracy after fine-tuning for 100 epochs.
\subsection{Ablation Studies on Each Component}
In this subsection, we present ablation studies on each component. 
Table~\ref{tab:ablation1} shows that our proposed teacher normalization achieves 0.3\% top-1 accuracy gain. Only inputting the masked tokens into the teacher network suffers from 0.5\% performance degradation due to insufficient information exploration. While using our proposed multi-fold masking strategy, we can achieve 0.6\% improvement compared with only masked token inputs. Besides, multi-fold masking even outperforms taking full image as inputs more efficiently. 
\begin{table}[!t]
\renewcommand{\baselinestretch}{1.0}
\renewcommand{\arraystretch}{1.0}
\setlength{\tabcolsep}{4pt}
\begin{center}
\caption{Ablation studies on each component. $``$Full image$"$ refers to the teacher network taking the whole image as input. $``$Only masked$"$ refers to only the masked patches that are fed into the teacher network. $``$Multi fold$"$ indicates our multi-fold masking strategy. Teacher normalization refers to whether adding feature values normalization within patches or not.}\label{tab:ablation1} 
\begin{tabular}{ccccc}
\hline \multirow{2}*{\shortstack{Full\\image}} & \multirow{2}*{\shortstack{Only\\ masked}} &\multirow{2}*{\shortstack{Multi\\fold}}  &\multirow{2}*{\shortstack{Teacher\\ normalize}}  & \multirow{2}*{Accuracy} \\
&&&&\\
\hline 
$\checkmark$ & $\times$ & $\times$ &$\times$ & $83.7$ \\
$\checkmark$ & $\times$ & $\times$ &$\checkmark$ & $84.0$ \\
$\times$ & $\checkmark$ & $\times$ &$\checkmark$ & $83.5$ \\
$\times$ & $\times$ & $\checkmark$ &$\checkmark$ & $\textbf{84.1}$ \\
\hline 
\end{tabular}
\end{center}
\end{table}

\subsection{The EMA Strategy}
This experiment is conducted without the multi-fold masking strategy to evaluate the raw performance of the EMA strategy. Specifically, we have two settings for the EMA strategy: (1) update the parameters of the teacher branch with EMA for each training iteration. (2) update the parameters of the teacher branch with EMA for each training epoch. As shown in \figurename~\ref{fig:ema} (a), conducting the EMA strategy to update the teacher branch each training iteration is extremely sensitive to the value of the momentum coefficient. Specifically, only changing the value by 0.001 results in the sharply degraded performance. In contrast, the EMA strategy to update the teacher branch per batch is more robust to the momentum coefficient. Besides, conducting the EMA strategy per epoch can better benefit from long training epochs.

\subsection{The Multi-fold Masking Strategy}
\begin{figure}[!t]
\centering
\includegraphics[width=12.2cm]{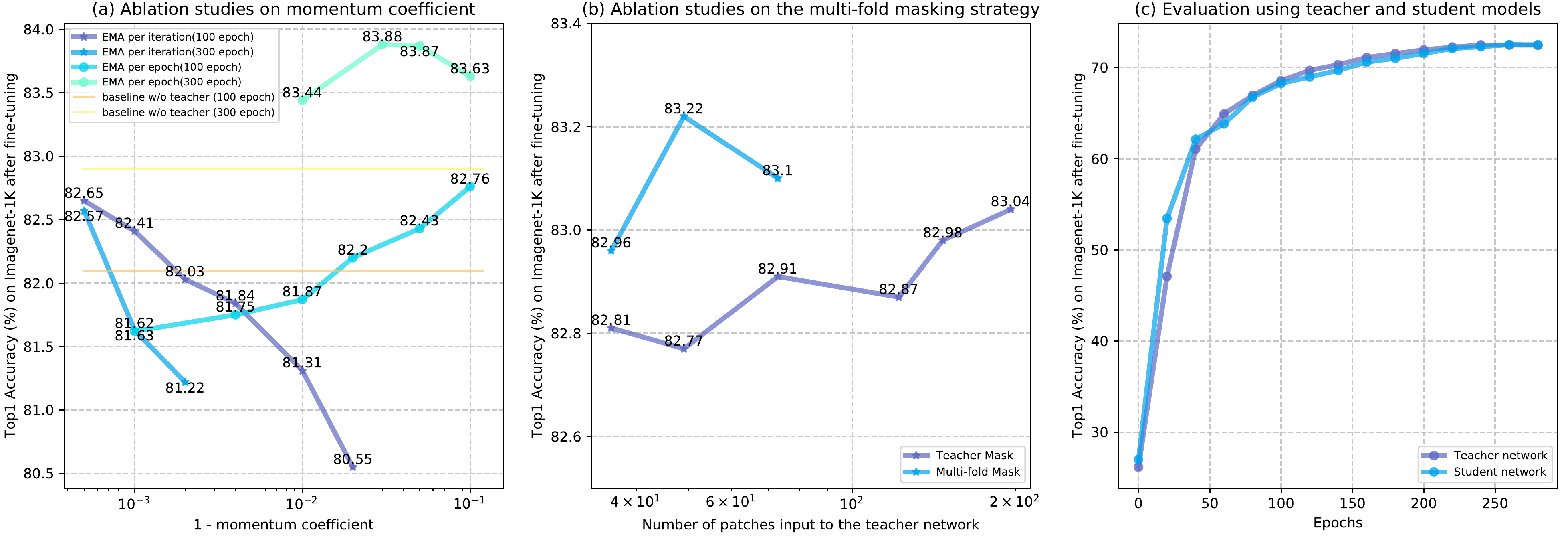}
\caption{(a) Ablation studies on the EMA update strategy for the coefficient of the momentum teacher. (b)Ablation studies on the multi-fold masking strategy. (c) Ablation studies on the evaluation of teacher and student models.}
\label{fig:ema}
\end{figure}

We further conduct an ablation study on the multi-fold masking strategy. 
Firstly, we study experiments on the teacher crop that mask some of the target tokens and feed the remaining tokens into the teacher. The whole image is divided into 196 image patches with the size of $16\times16$, and we randomly sample a different number of total patches, \textit{e.g.}, 36, 49, 79, 122, 147 and 196 as whole image input into the teacher network. 
As shown in \figurename~\ref{fig:ema} (b), for the case that 36 image patches (roughly 18\% of the whole image) are input to the teacher network, our method can achieve 82.81\% top-1 accuracy. 83.04\% top-1 accuracy is achieved when we take the whole image as input. Comparably, only 0.23\% performance gain is achieved when five times as many image patches are input. The problem of spatial redundancy is also common in the input views of the teacher network.

Correspondingly, we take the multi-fold masking strategy. Apart from the 49 patches that are input to the student network, the left 147 image patches are divided into 2 fold as 2$\times79$ patches, 3 fold as 3$\times49$ patches or 4 fold as 4$\times36$ patches. As shown in \figurename~\ref{fig:ema} (b), a multi-fold masking strategy can consistently improve the performance. For 3 fold with 49 patches, our method achieves the best performance. This experiment also proves our information bottleneck discussion, that with comparable mutual information between each fold and the student input, we can get the best performance.

\subsection{Evaluation of Teacher and Student Models}
As shown in \figurename~\ref{fig:ema} (c), for every 20 epochs of our training, we finetune our pre-trained teacher and student models for 5 epochs on ImageNet-1K. The student model performs better than the teacher model in the initial epochs when the learning rate is relatively low due to warm up and then is very close to the teacher when the learning rate increases. 
We demonstrate that the teacher model and student model trained by our SdAE achieve similar performance. That proves the teacher branch can really learn high semantic level related representations. The teacher model can be seen as an ensemble over many student models. We follow most previous works to use the student model as the final model. 

\section{Related Work}
Previous self-supervised methods utilize different priors of images to design clever pretext tasks, such as predicting the patch positions~\cite{doersch2015unsupervised}, inpainting~\cite{pathak2016context}, colorization~\cite{zhang2016colorful}, and rotation prediction~\cite{gidaris2018unsupervised}. Recent progress in self-supervised learning focuses on contrastive learning and masked image modeling.

\textbf{Masked Image Modeling.} 
Motivated by BERT~\cite{bert2018} for masked
language modeling(MLM), masked image modeling(MIM) is proposed to learn representations from images corrupted by masking. Recent works leverage the powerful Vision Transformer, which matches the masked image modeling task. iGPT~\cite{chen2020generative} operates on sequences of pixels and predicts unknown pixels. Vision transformer~\cite{AlexeyDosovitskiy2021ViT} studies masked average value prediction for self-supervised learning. BEiT~\cite{HangboBao2021BEiT} proposes to predict discrete tokens based on a pre-trained image tokenizer while iBOT~\cite{JinghaoZhou2021iBOT} proposes an online tokenizer. MAE~\cite{he2021masked} proposes a masked autoencoder for reconstructing the image pixels. 
CAE~\cite{XiaokangChen2022CAE} provides a two-branch network for MIM, and the masked features are modeled as a regularization to align the mask and unmasked features.

\textbf{Contrastive Learning.} Contrastive learning is proposed based on the InfoMax principle, which aims at maximizing the mutual information~\cite{wu2018unsupervised} across different augmentations of the same image~\cite{chen2020simple,tian2020contrastive}. 
This augmentation invariance is achieved by enforcing the similarity over different views of the same image while avoiding model collapse. Model collapse can be avoided by introducing negative samples for noise-contrastive estimation~\cite{TingChen2020BigSM,chen2020simple,he2020momentum,wu2018unsupervised}. The models typically regard various data augmentations as different views of an image and then make the representations of positive pairs similar while pushing negative pairs away. Large memory banks or large batch size is leveraged to obtain more informative negative samples. BYOL~\cite{JeanBastienGrill2020BYOL} and SimSiam~\cite{chen2021exploring} employ an asymmetric network and eliminate the requirement of negative samples. Other methods use clustering~\cite{asano2019self,caron2018deep,caron2020unsupervised,caron2021emerging} to organize image examples.

\section{Conclusion}
In this paper, we propose a novel Self-distillated Masked Autoencoder for masked feature reconstruction, namely SdAE.  
We first formulate the framework of the masked image modeling task, based on which we analyze that existing methods are sub-optimal due to the need of pre-trained codebooks or just reconstructing the low-level pixels. 
We propose a self-distillated framework for reconstruction in the high-level feature space. Besides, we analyze how to build good views for the teacher branch to produce latent representation from the perspective of information bottleneck theory and propose a multi-fold masking strategy that can also relieve the spatial redundancy. Experimentally, based on a vanilla ViT-Base model, our SdAE achieves a new state-of-the-art of 84.1\% top-1 accuracy with only 300 epochs pre-training.

\noindent \textbf{Acknowledgments.} This work was supported in part by the National Natural Science Foundation of China under Grant 61932022, Grant 61931023, Grant 61971285, Grant 62120106007, and in part by the Program of Shanghai Science and Technology Innovation Project under Grant 20511100100.




%
%
\clearpage

\appendix

\section{Appendix}

\newcommand{\lr}{\emph{lr}\xspace}
\newcommand{\wtd}{\emph{wd}\xspace}
\newcommand{\drp}{\emph{dp}\xspace}
\newcommand{\expnum}[2]{{#1}\mathrm{e}^{#2}}

\subsection{Implementation Details}

\textbf{Pretraining.} The settings are almost the same as MAE~\cite{he2021masked}. We use AdamW for optimization and train the CAE for 300 epochs with batch size 768. We set the learning rate as 8e-4, with cosine learning rate decay and a 60 epoch warmup, and set the weight decay as 0.05. We employ the drop path with the ratio of 0.25 only on the encoder. The momentum coefficient is set as 0.96 and with a cosine schedule to 0.99, and EMA is conducted per pre-training epoch. We set the mask ratio as 0.75 with only 49 tokens fed into the student branch. The masked tokens are divided into 3 folds where each fold contains 49 tokens, and all folds are fed into a shared weighted teacher branch.

\textbf{Fine-tuning on ImageNet.} We follow the fine-tuning setting almost the same as MAE~\cite{he2021masked} to use layer-wise learning rate decay, weight decay, and AdamW. The batch size is 2048, and the weight decay is 0.05. For ViT-B, we train 100 epochs with base learning rate 5e-4, layer-wise decay rate 0.65, drop path rate 0.1, and warmup epoch 10. For ViT-L, we train 50 epochs with base learning rate 1e-3, layer-wise decay rate 0.75, drop path rate 0.2, and warmup epoch 5.

\textbf{Object Detection and Instance Segmentation on COCO.} We utilize the same setting as CAE~\cite{XiaokangChen2022CAE} that uses multi-scale training and resizes the image with the size of the short side between 480 and 800 and the long side no larger than 1333. The batch size is 32, the learning rate is 3e-4, and the layer-wise decay rate is 0.75. We train the network with the 1$\times$ schedule: 12 epochs with the learning rate decayed by 10$\times$ at epochs 9 and 11. We do not use multi-scale testing. The Mask R-CNN implementation follows MMDetection~\cite{mmdetection}.

\textbf{Semantic Segmentation on ADE20K.} We utilize the same setting as CAE~\cite{XiaokangChen2022CAE}. We use AdamW as the optimizer. The batch size is 16 and the layer-wise decay rate is $0.65$. The input resolution is $512 \times 512$. We use the learning rates as 4e-4 for all the results in our experiments. We conduct fine-tuning for $160 \mathrm{~K}$ steps, and we do not use multi-scale testing.

\subsection{More Results for Larger Models and Longer Pre-training Epochs}
SdAE can also perform well with only 300 epochs pre-training on a larger model scale such as ViT-L.
We study the fine-tuning on the ILSVRC-2012 ImageNet dataset~\cite{russakovsky2015imagenet} with 1k classes and 1.3M images. For a fair comparison, we directly follow most of the hyperparameters of MAE~\cite{he2021masked} in our fine-tuning experiments. All reported experimental results are only fine-tuning for 50 epochs.

As shown in Table~\ref{tab:classify}, compared with the models trained by random initialization (train from scratch), our pre-trained SdAE significantly improves the performance. 
Specifically, vision transformers trained from scratch only achieve 82.6\% top-1 accuracy with  ViT-L. While our SdAE achieves 85.7\%, demonstrating the effectiveness of pre-training with unlabeled data. 

Compared with previous self-supervised methods for vision transformers, our proposed SdAE surpasses them on ImageNet fine-tuning by a large margin. For ViT-L, our SdAE outperforms
MoCo v3 by 1.6\% top-1 accuracy with the same number of training epochs and our SdAE outperforms MAE by 1.4\% top-1 accuracy with the less number of training epochs. Besides, our SdAE outperforms BEiT by 0.5\% top-1 accuracy, while BEiT requires an additional pre-trained codebook and longer training epochs. In addition, our SdAE outperforms iBOT by 0.7\% top-1 accuracy. Moreover, compared to the 1600 epoch pre-trained ViT-L of MAE and MaskFeat, which requires very large computational costs, our SdAE can achieve comparable performance.
\begin{table}[!t]
\renewcommand{\baselinestretch}{1.0}
\renewcommand{\arraystretch}{1.0}
\setlength{\tabcolsep}{6pt}
\begin{center}
\caption{Image classification results on the ILSVRC-2012 ImageNet dataset with top-1 accuracy. $``$Epochs$"$ refers to the number of pre-training epochs. MoCo v3 adopts multi-crop augmentation with 2 global crops of 224 × 224 for pre-training.}\label{tab:classify} 
\begin{tabular}{lcccc}
\hline Method & Epochs & Crops & Accuracy \\
\hline \emph{Methods using ViT-L}: & &   \\
Train from Scratch&  $300$ & $-$ & 82.6  \\
MoCo v3 & 300 & 2 & $84.1$   \\
BEiT & 1600 & 1 & $85.2$   \\
iBOT & 250 & 1 & $85.0$   \\
MAE & 400 & 1 & $84.3$  \\
MAE & 1600 & 1 & $\textbf{85.9}$  \\
MaskFeat & 1600 & 1  & $85.7$ \\
SdAE & 300 & 1 & $85.7$ \\
\hline
\end{tabular}
\end{center}
\end{table}

\begin{table}[!t]
\renewcommand{\baselinestretch}{1.0}
\renewcommand{\arraystretch}{1.0}
\setlength{\tabcolsep}{6.5pt}
\begin{center}
\caption{Longer pre-training epochs results with overall computational costs and memory cost compared with MAE, SimMIM and CAE on ViT-Base using NVIDIA V100 GPUs. Ft: Image classification on the ImageNet dataset with top-1 accuracy under fine-tuning 100 epochs. Det: object detection on the COCO dataset, and AP$^b$ is reported. Seg: semantic segmentation on the ADE20K dataset, and mIoU is reported.}\label{tab:timecompare}
\begin{tabular}{ccccccc}
\hline
Method& Epoch & Train Time& GPU Mem& Ft& Det & Seg \\
\hline
MAE & 300& 45.0h & 14.93G& 82.9& 45.4  & 45.8\\
CAE & 300 & - & - & 83.3 & 48.0 & 47.7\\
SdAE & 300 & 60.8h  & 16.02G& \textbf{84.1}& \textbf{48.9} & \textbf{48.6}\\
\hline
SimMIM & 800 & 188.6h &20.32G &83.8 &46.5&46.9\\
MAE & 800& 140.1h & 14.93G&83.4&47.8  &47.3\\
CAE & 800 & - & - & 83.6 & 49.2 & 48.8\\
MAE & 1600& 278.5h &14.93G &83.6& 48.4 & 48.1\\
SdAE & 800 & 174.1h  & 16.02G& \textbf{84.0}& \textbf{49.7} & \textbf{49.0}\\

\hline
\end{tabular}
\end{center}
\end{table}
As shown in Table~\ref{tab:timecompare}, although the cost of SdAE \emph{de facto} surpasses MAE per epoch, it can speed up convergence and achieve comparable performance in fewer epochs. It is also more efficient than SimMIM that adopts the mask token for the encoder.
For longer pre-training epochs, SdAE is faced with a little bit performance degradation on ImageNet fine-tuning. We speculate that this is due to the fact that the Vit-base capacity is relatively close to the performance upper bound of the MIM tasks.
However, SdAE shows continuous performance enhancement with longer training epochs on ADE20K semantic segmentation and COCO object detection, which also surpasses other methods by a considerable margin.

\subsection{Comparison of iBOT, data2vec, and SplitMask}
\begin{figure}[!t]
\centering
\includegraphics[width=12.1cm]{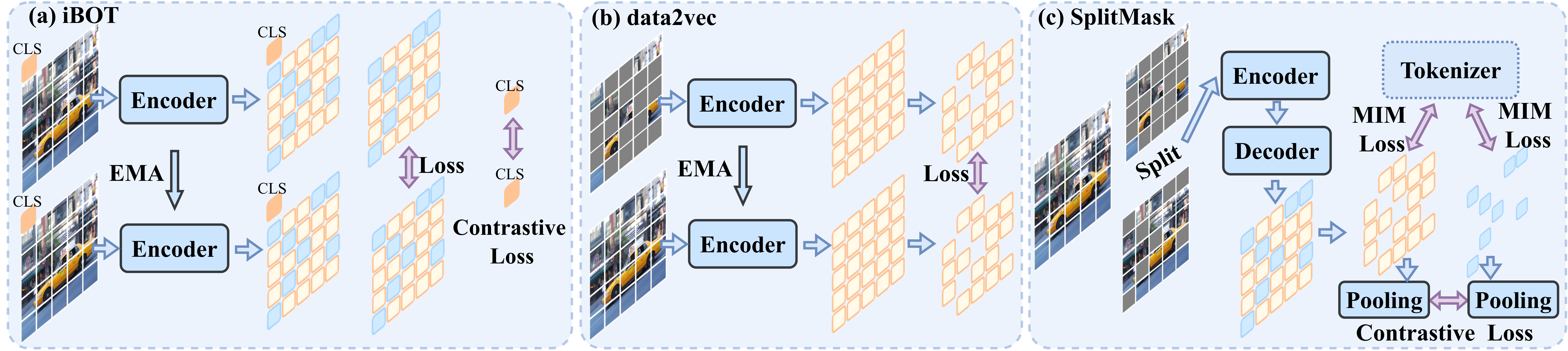}
\caption{
Comparison of other recently proposed generative-based self-supervised learning methods. (a) iBOT heavily relies on contrastive loss. (b) data2vec fails to consider the redundancy existing in the mask and visible tokens. (c) SplitMask fails to consider using a two-branch distillation structure and still needs an extra tokenizer.}
\label{fig:architecture2}
\end{figure}

Except for the comparison of typical generative-based self-supervised learning methods in \figurename~\ref{fig:architecture2} such as BeiT~\cite{HangboBao2021BEiT}, PeCo~\cite{XiaoyiDong2021PeCo}, MAE~\cite{he2021masked} and CAE~\cite{XiaokangChen2022CAE}, we also provide the comparison of several recently proposed works in \figurename~\ref{fig:architecture2}.

iBOT~\cite{JinghaoZhou2021iBOT} is more likely a contrastive learning/instance discrimination-based method. iBOT needs careful parameter setting of multi-crop augmentation, which uses 10 local crops with local scale being (0.05,0.32) and global scale being (0.32,1.0). In addition,
iBOT heavily depends on the contrastive loss.
MIM without the class token contrastive loss leads to undesirable results of $9.5 \%$ kNN accuracy and $29.8\%$ linear accuracy on iBOT, indicating that iBOT benefits more from contrastive structure than MIM to extract the visual semantics.

Data2vec~\cite{baevski2022data2vec} also uses an EMA parameterization of the two-branch teacher-student distillation structure. However, data2vec does not consider the spatial redundancy existing in the network structure. Not only visible unmasked patches but also learned mask embedding tokens are fed into the student branch. Furthermore, the whole input image will be fed into the teacher branch, ignoring the reconstruction loss computed only on masked tokens, which will also increase computational costs.
In addition, data2vec conducts an EMA update per iteration. The MIM output and reconstruction target will be very similar if the momentum coefficient is not extremely small. So the network is sharply sensitive to the momentum coefficient. So, data2vec needs precisely tuning on this coefficient where in ViT-L they need to first set momentum coefficient as 0.9998 for the first 800 epochs and then reset the learning rate schedule and the teacher weights to the student and continue for another 800 epochs with momentum coefficient as 0.9999.

SplitMask~\cite{El2021SplitMask} considers the redundancy of split tokens. However, SplitMask still needs an additional tokenizer to produce discrete latent representations to conduct MIM. Moreover, SplitMask does not consider using a two-branch network to distill the representation between split tokens but adds a pooling module to calculate the contrastive loss between global representations.

\begin{figure}[!t]
\centering
\includegraphics[width=7cm]{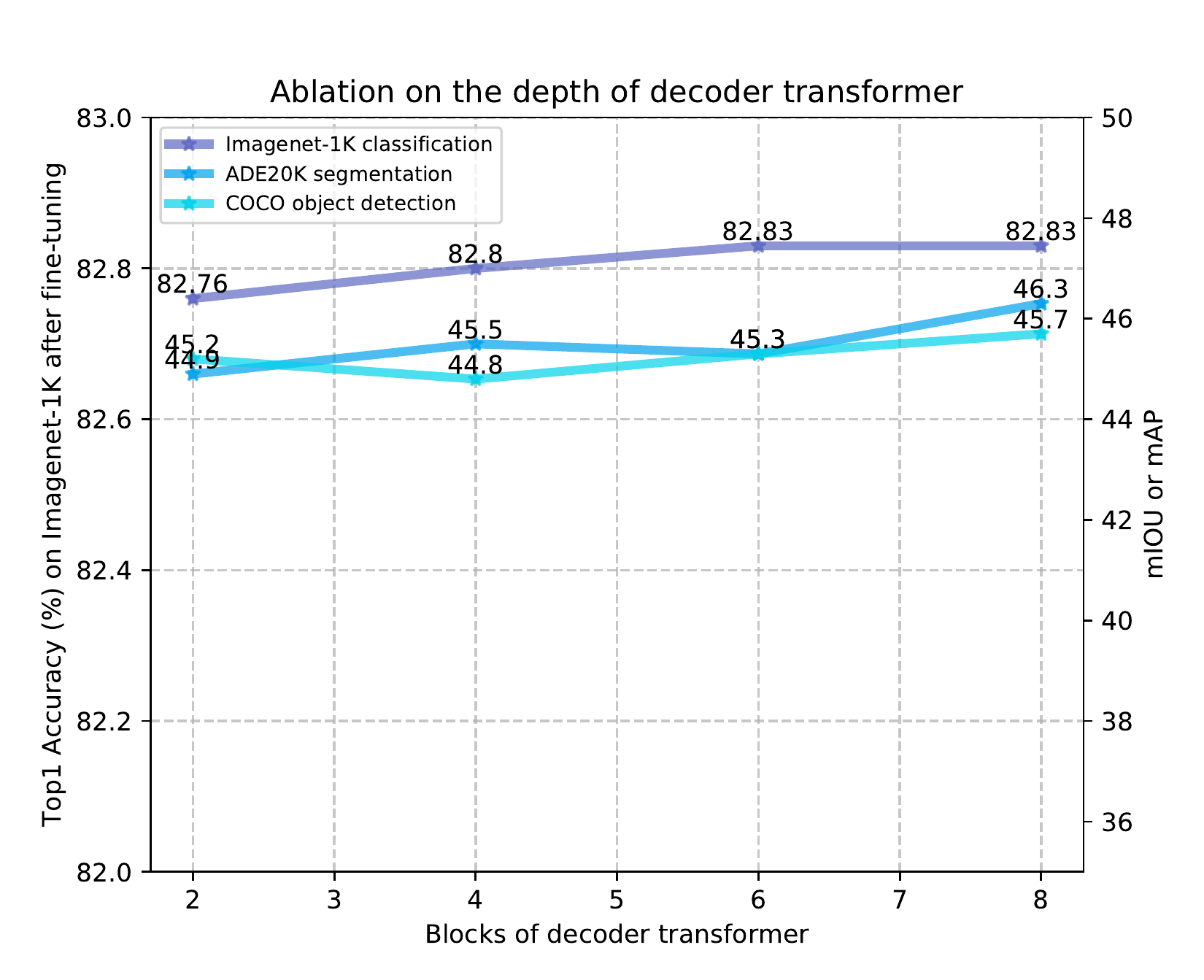}
\caption{Ablation studies on the depth of the decoder transformer.}
\label{fig:decoder}
\end{figure}

\begin{figure}[!t]
\centering
\includegraphics[width=12.0cm]{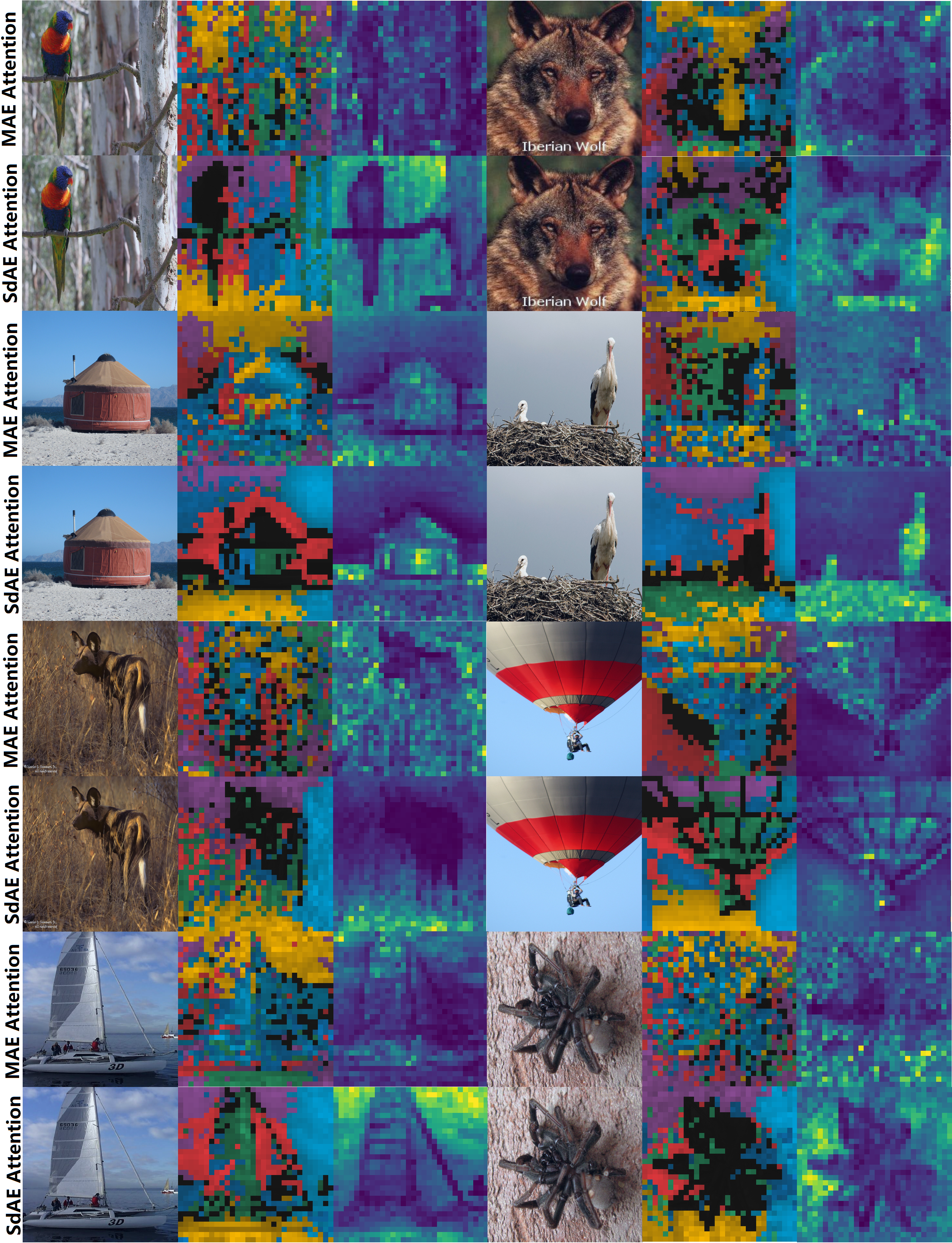}
\caption{Attention maps visualization of MAE and SdAE. The first column and forth column place the original images. 
The second column and fifth column are the visualization of different heads from the last layer with different colors. 
The third column and sixth column are the visualization of mean of all attention heads. 
}
\label{fig:attn_map}
\end{figure}

\subsection{Ablations on the Depth of Decoder}
The decoder of the autoencoder, which maps the latent representation back to the reconstruction space, plays an essential role in the masked image modeling task. In the language MLM, the decoder predicts missing words that contain rich semantic information so that the decoder can be trivial (an MLP) in BERT~\cite{bert2018}. However, in MAE~\cite{he2021masked}, the decoder reconstructs the image pixels, which reconstructs the latent representations into the low-level pixel space. Thus, MAE requires a relatively powerful decoder. In comparison, our student network maps the latent representations to the high-level semantic features so that the decoder can be lighter. That is another potential advantage of SdAE. As shown in \figurename~\ref{fig:decoder}, the experiment shows that the depth of the decoder has little impact on the performance. Specifically, even with two layers of the decoder transformer, our SdAE achieves 82.76\% top-1 accuracy and 45.2 mAP on COCO object detection, which only suffers 0.07\% and 0.5 mAP performance degradation compared with eight layers of the decoder transformer.

\subsection{Visualization}
To analyze, we visualize the self-attention map with 300-epoch pre-trained ViT-B/16 of both MAE and SdAE. We choose the class token as the query and visualize attention maps from different heads of the last layer with different colors, following iBOT~\cite{JinghaoZhou2021iBOT}. As shown in \figurename~\ref{fig:attn_map}, we indicate that SdAE shows the capability to learn high-level semantic features to separate different parts of objects. Compared with MAE, SdAE is able to learn more meaningful high semantic information.

Specifically, in the figure, we observe SdAE can distinguish the bird from the tree or distinguish the eyes and ears of the Iberian wolf. Moreover, SdAE can also focus on the discriminative details of the object (e.g., the skeleton of a hot air balloon and sailboat rope) without using the contrastive loss. For more complex scenes like spiders on the surface of complex texture SdAE is still able to distinguish subjects. This is because SdAE does not need to reconstruct every pixel, so it did not pay attention to useless details. 

With only a simple normalized feature MSE loss, we can achieve similar behaviors with intricately designed instance discrimination methods such as DINO~\cite{caron2021emerging}.

\bibliographystyle{splncs04}
\bibliography{egbib}

\end{document}